\documentclass[graybox]{svmult}
\usepackage[first=0,last=9]{lcg}
\usepackage{scalerel}   

\usepackage{wrapfig}

\usepackage{mathptmx}       
\usepackage{helvet}         
\usepackage{courier}        
\usepackage{makeidx}         
\usepackage{graphicx}        
\usepackage{multicol}        
\usepackage{wrapfig}

\makeindex             

\usepackage{amssymb}
\usepackage{amsmath}
\usepackage{algorithm}
\usepackage{algorithmic}

\usepackage{multirow}
\usepackage{array}
\usepackage{tikz}           
\usepackage{color, colortbl}
\usepackage{makecell}

\definecolor{Gray}{gray}{0.9}

\begin{document}

\title*{Integrating Product Coefficients for Improved 3D LiDAR Data Classification}

\author{Patricia Medina}

\institute{
Patricia  Medina \at New York City College of Technology, CUNY;\\ \email{patriciamg90@gmail.com}
 }

\maketitle


\abstract{In this paper, we address the enhancement of classification accuracy for 3D point cloud Lidar data, an optical remote sensing technique that estimates the three-dimensional coordinates of a given terrain. Our approach introduces product coefficients, theoretical quantities derived from measure theory, as additional features in the classification process. We define and present the formulation of these product coefficients and conduct a comparative study, using them alongside principal component analysis (PCA) as feature inputs. Results demonstrate that incorporating product coefficients into the feature set significantly improves classification accuracy within this new framework.}

\section{Introduction}
%

LiDAR technology, which employs laser beams to measure distances, integrates GPS and inertial navigation data to generate precise 3D point representations of reflective surfaces. The return time of the laser beam determines distance measurements between the sensor and target, facilitating the creation of detailed 3D maps. See \cite{inbook} for a technical treatment in remote sensing.

These 3D LiDAR point clouds serve numerous applications across geosciences, including updating digital elevation models, monitoring glacial changes and landslides, shoreline mapping, and assessing urban development. A crucial step in these applications is the classification of point cloud data into fundamental categories such as vegetation, structures, and water. Depending on the analysis requirements, classification schemes may range from binary segmentation (e.g., ground vs. non-ground) to more granular surface-based distinctions (e.g., gravel, sand, rock in reservoir mapping). The development of classification algorithms, particularly those leveraging multi-scale intrinsic dimensionality approaches, continues to be an area of active research \cite{BRODU,BaIzMcNeShships}.

Our study explores classification frameworks for 3D LiDAR point clouds by investigating feature engineering techniques and dimensionality reduction. Specifically, we augment raw point data with product coefficients computed within each point’s local neighborhood and apply Principal Component Analysis (PCA) to enhance feature representations. Experimental results demonstrate that these modifications significantly improve classification performance, as reflected in higher F1 scores, increased accuracy, and reduced error rates.

Rather than classifying entire shapes, our approach focuses on point-wise classification. To systematically evaluate different algorithmic combinations, we develop a preliminary framework that applies feature transformation techniques followed by classification models. Our methodology involves engineering new features (such as product coefficients), applying linear dimensionality reduction (PCA), and training two classification algorithms: Random Forest and k-Nearest Neighbors. We have done feature engeneering in \cite{Medina2021} by using what we called the ``neighbour matrix'' in a similar dataset. Our new approach includes now the product coefficients as new locally generate features, hence taking advantage of them as quantities that have been used for decision rules. 

While PCA has been effective in reducing feature covariance and improving classifier performance, alternative dimensionality reduction techniques could yield comparable or superior results. Future work will explore such alternatives to optimize classification accuracy further.

LiDAR data collection platforms vary widely, including aircraft, helicopters, ground vehicles, and tripods. This flexibility enables applications such as fault detection, forest inventory analysis, beach volume assessments, landslide risk evaluation, and habitat mapping \cite{Moskal}. ArcGIS provides tools for managing, visualizing, analyzing, and sharing the resulting dense point cloud datasets \cite{arcgis}.

The paper is structured as follows: Section \ref{sec:data} describes the attributes of LiDAR data. In Section \ref{PCs}, we introduce product coefficients and their role in representing measures, detailing their computation in dyadic sets (binary trees). Section \ref{subsec: PCs on lidar} provides a focused discussion on computing product coefficients within a 3D point cloud embedded in the unit cube. Section \ref{subsec: PCA and PCs experients} presents
 the classification frameworks, describes experimental setups, and summarizes classification performance, with results shown in Tables \ref{table:F1scoresNoPCvsPC} and \ref{table:F1scoresPCsPCA}. Finally, Section \ref{summary} concludes with key findings and potential future research directions.

\section{The Data}\label{sec:data}

LiDAR point clouds can be categorized into various classes, such as terrain, vegetation, buildings, and water surfaces. These classifications are encoded using integer values in LAS files, following the standards established by the American Society for Photogrammetry and Remote Sensing (ASPRS). The latest LAS specification defines eighteen distinct classes, including 0 (Never Classified), 1 (Unassigned), 2 (Ground), 3 (Low Vegetation), 4 (Medium Vegetation), 5 (High Vegetation), 6 (Buildings), 7 (Noise), 9 (Water), 10 (Rail), and 17 (Bridge Deck), among others.

\begin{figure}
	\centering	
	\includegraphics[scale=0.7]{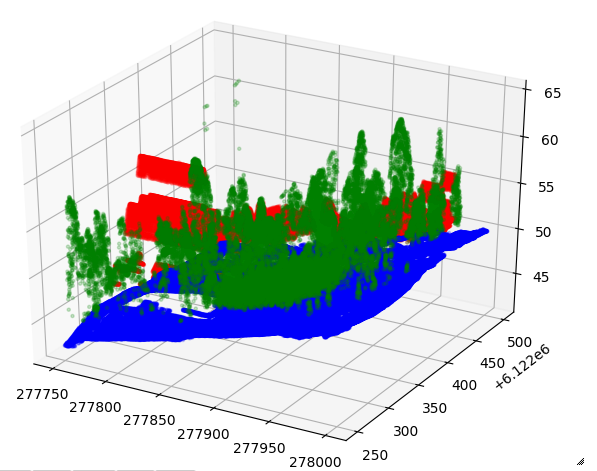}
	\caption{Visualization of a LiDAR point cloud from an urban neighborhood in Australia, containing 277,572 points. The dataset includes multiple classes, with the most visually prominent ones being terrain, structures, and vegetation.}	
	\label{fig:scenario}
\end{figure}

For this study, we utilize a publicly available LiDAR dataset, originally introduced in LAS tool tutorials several years ago. The dataset contains 277,572 points and is labeled with five classification categories: ground, high vegetation, low vegetation, buildings, and water. A Python-generated visualization of the dataset, color-coded by class, is presented in Fig. \ref{fig:scenario}. Notably, the dominant categories in the visualization are ground, buildings, and high vegetation.  

\begin{figure}\label{fig:tree}
	\centering	
	\includegraphics[scale=0.8]{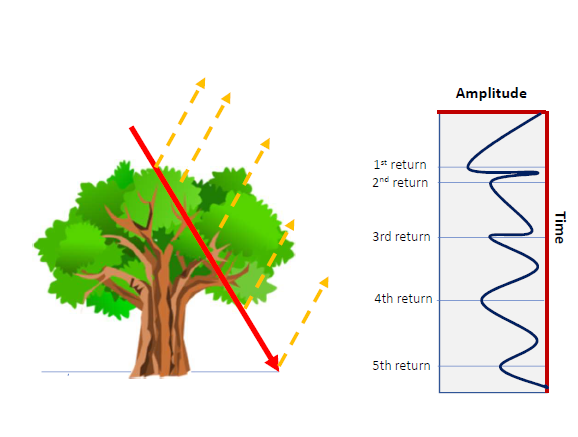}
	\caption{\scriptsize An illustration of LiDAR pulse returns interacting with different surfaces. A single pulse may reflect off multiple surfaces, such as tree branches and the ground, producing multiple returns. This characteristic is essential for analyzing vertical structures in the dataset. Adapted from \cite{medina2019heuristic}. }
\end{figure}

Each LiDAR point is associated with multiple attributes that provide valuable information about the scanned environment. The key attributes include:  

\begin{description}

\item[Intensity:] Measures the strength of the returned laser pulse, influenced by the reflectivity of the surface. This attribute helps differentiate between surface materials, such as distinguishing between vegetation and pavement.  

\item[Return Number:]Indicates the sequential order of multiple returns from a single emitted pulse. A pulse that interacts with multiple surfaces (e.g., tree canopy and ground) will generate multiple returns, each assigned a return number (first return, second return, etc.).  

\item[Number of Returns:] Represents the total number of reflections recorded from a single laser pulse. For example, pulses penetrating dense vegetation often produce multiple returns, while those hitting solid surfaces typically generate a single return. An illustration of this process is shown in Fig. \ref{fig:tree}.  

\item[Point Classification:] Assigns a category to each LiDAR point, using standardized numerical codes. These classifications enable applications such as terrain modeling, building detection, and hydrological analysis.  

\item[Edge of Flight Line:] A binary indicator (0 or 1) that identifies whether a point lies at the boundary of a scanning pass. This flag is useful for assessing variations in data quality along flight edges.  

\item[RGB Values:] Red, Green, and Blue color information, typically derived from co-registered aerial imagery. This enhances visualization and aids in classification tasks, such as distinguishing urban features from natural landscapes.  

\item[GPS Time:] Records the timestamp (in GPS seconds of the week) when the laser pulse was emitted, ensuring accurate geospatial alignment of the collected points.  

\item[Scan Angle:] Defines the angle at which the laser pulse was emitted relative to the nadir (directly downward). Angles range from -90 to +90 degrees, with 0° corresponding to a nadir shot.  

\item[Scan Direction:] Indicates whether the scanning mirror was moving in a positive (left to right) or negative (right to left) direction at the moment of pulse emission. This attribute helps in scan alignment and data correction.  

\end{description}

In this study, we focus on a minimal feature set for classification. Specifically, we retain only the spatial coordinates \((x, y, z)\) and generate additional features using product coefficients derived from these three spatial dimensions. By applying these transformations, we aim to explore how feature engineering influences classification accuracy while maintaining a simple yet effective approach.


\section{The product coefficients}\label{PCs}

In this section, we formally define product coefficients and outline their computation using a binary tree derived from a given set \(X\). Many datasets can be represented as sequences of vectorized observations, which naturally correspond to discrete metric-measure spaces. These structures allow for the systematic derivation of multi-scale representations, whose existence is guaranteed by established mathematical results.  

A key advantage of these canonical representations is their flexibility—they are independent of any specific domain, making them applicable across diverse datasets. This property enables them to be leveraged for pattern recognition, anomaly detection, and automated decision-making in machine learning applications. Since these representations capture fundamental data characteristics at multiple scales, they facilitate a structured approach to identifying deviations from expected behavior.  

To construct these representations, we rely on dyadic sets, which are collections of subsets organized into a hierarchical binary tree structure. Dyadic sets arise naturally in various contexts, such as partitioning the unit interval, feature spaces, or unit cubes. More formally, we consider a dyadic set \(X\), where elements are recursively subdivided into smaller subsets, forming an ordered binary tree.

We first recall that a dyadic set is a collection of subsets structured as an ordered binary tree (e.g. unit interval, feature sets, unit cubes). More precisely, we consider a {\it dyadic} set $X$ 
which is the {\it parent} set or root of the ancestor tree of a system of left and right {\it child} subsets. For each subset $S$ (dyadic subset) of $X$, we denote the left child by $L(S)$ and the right child by $R(S).$  Let $\mu$ be a non-negative measure on $X$ and $dy$ the naive measure, such that $dy(X)=1$.  
\begin{equation*}
	dy(L(S))=\frac{1}{2}dy(S),\quad dy(R(S))=\frac{1}{2}dy(S)
\end{equation*}

Note that $\mu$ is additive in the binary set system, i.e. $\mu (S)= \mu(L(S) \cup R(S))=\mu(L(S))+ \mu(R(S))$ ($L(S)$ and $R(S)$ are disjoint.)

\begin{figure}
\centering	
\includegraphics[trim={7cm 12cm 10cm 1cm},clip,scale=0.6]{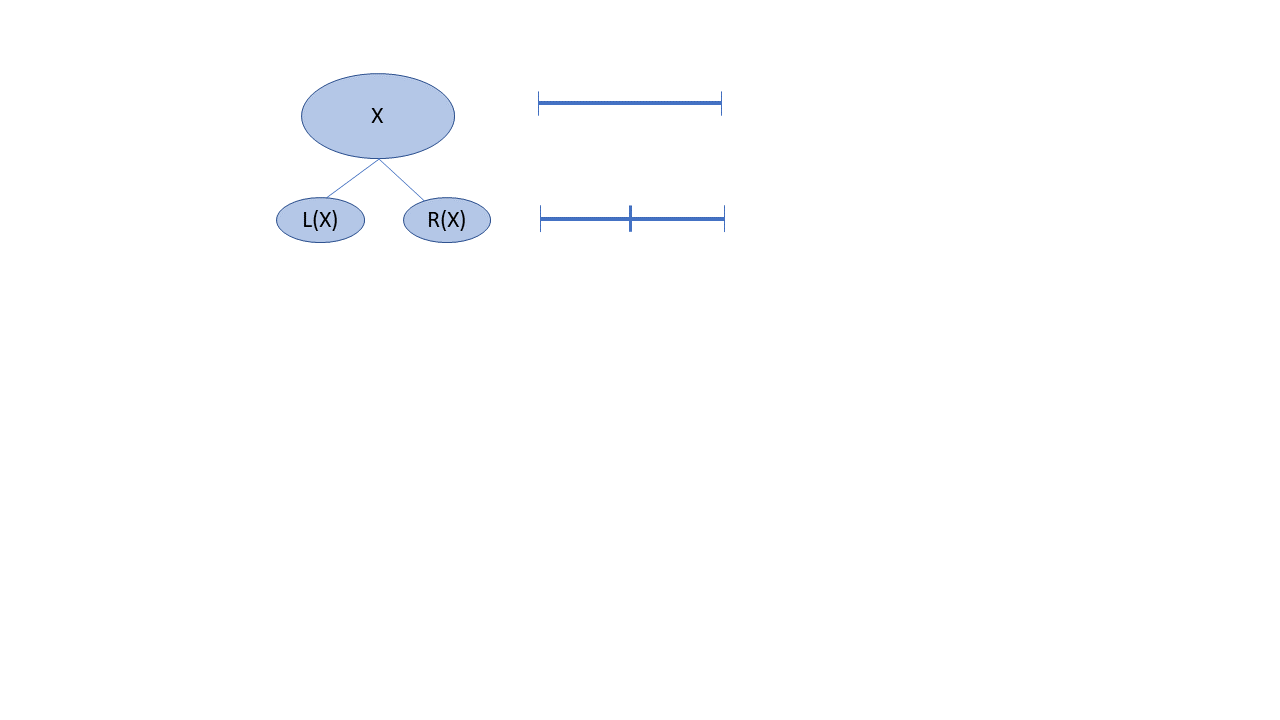}
\caption{illustration of the first level of a dyadic set or binary tree for a set $X$. On the right we have how this translate into an interval}
\label{fig:dyadic}
\end{figure}


$\mu$: non-negative measure on $X$; 
On Fig.\,\ref{fig:dyadic}, we can consider $dy$ the naive measure, such that $dy(X)=1$ if we work for instance in the interval $[0,1].$
\begin{equation*}
	dy(L(S))=\frac{1}{2}dy(S),\quad dy(R(S))=\frac{1}{2}dy(S)
\end{equation*}
$\mu$ is additive in the binary set system,
 i.e. $\mu (S)= \mu(L(S) \cup R(S))=\mu(L(S))+ \mu(R(S))$ ($L(S)$ and $R(S)$ are disjoint.)

Let $\mu$ be a dyadic measure on a dyadic set $X$ and $S$ be a subset of $X$. The {\it product coefficient} parameter $a_S$  is the solution for the following system of equations
\begin{eqnarray}
	\mu(L(S)) &=& \frac{1}{2}(1 + a_s) \mu(S) \label{eqn:1}\\
	\mu(R(S)) &=& \frac{1}{2}(1 - a_s) \mu(S)  \label{eqn:2}
\end{eqnarray}	


\begin{definition}
	Let $\mu$ be a dyadic measure on a dyadic set $X$ and $S$ be a subset of $X$. The {\it product coefficient} parameter $a_S$  is the solution for the following system of equations
	\begin{eqnarray}
		\mu(L(S)) &=& \frac{1}{2}(1 + a_s) \mu(S) \label{eqn:1}\\
		\mu(R(S)) &=& \frac{1}{2}(1 - a_s) \mu(S)  \label{eqn:2}
	\end{eqnarray}	
\end{definition}
A solution for \eqref{eqn:1}--\eqref{eqn:2} is unique if $\mu(S) \neq 0.$ If $\mu(S)=0$, we assign the zero value to the product coefficient, i.e., $a_S=0.$
Note that if $\mu(S)>0$ then solving  \eqref{eqn:1}--\eqref{eqn:2} for $a_s$ gives 
\begin{equation}
	a_s=\dfrac{\mu(L(S)) - \mu(R(S)) }{\mu(S)}
\end{equation}


The product coefficients are bounded, $|a_S| \leq 1.$  In what follows, we use a Haar-like function $h_S$ defined as 

\begin{equation}\label{Haar-like}
	1 \mbox{ on } L(S),\, -1 \mbox{ on } R(S), \mbox{ and } 0 \mbox{ on } X-S.
\end{equation}	

	
\begin{example}[Formula for a scale 0 dyadic measure]
	Let $X=[0,1]$ and let there be a non-negative measure $\mu$ such that $\mu(X)=1,\, \mu(L(X))=\frac{1}{4}$ and $\mu(R(X))=\frac{3}{4}.$ Let $a=a_X$ be the product coefficient which is the solution for the system of equations
	
	\begin{eqnarray}
		\mu(L(X)) &=& \frac{1}{2}(1 + a) \mu(X) \label{eqn:1X}\\
		\mu(R(X)) &=& \frac{1}{2}(1 - a) \mu(X)  \label{eqn:2X}.
	\end{eqnarray}	
	
	Subtracting \eqref{eqn:2X} from \eqref{eqn:1X} we obtain $a=\dfrac{\mu(L(X))-\mu(R(X))}{\mu(X)}=-\dfrac{1}{2}.$

\end{example}

Since, $dy(X)=1$ and $dy(L(X))=\frac{1}{2}=dy(R(X))$ then by the product formula form,
\begin{equation}
	\mu=\mu(X)(1+ah)dy,
\end{equation}
where $h$ is the Haar-like function with $S=X.$ 

In this paper, we use the counting measure on Lidar data points instead of Borel measure on intervals. We compute product coefficients in scale 0 (only one coefficient), scale 1 (two coefficients) and scale 2 (3 coefficients.)

The product formula for non-negative measures in $X=[0,1]$ using the product factors $a_S$ first appeared in \cite{Fefferman}. We present the representation lemma for dyadic sets extracted from \cite{Linda}.


\begin{lemma}[Dyadic Product Formula Representation]\label{lemma1}

	Let $X$ be a dyadic set with binary set system $B$ whose non-leaf sets are $B_n$.
	
	\begin{enumerate}
		\item A non-negative measure $\mu$ on $X$ has a unique product formula representation 
		\begin{equation}
			\mu=\mu(X) \prod_{S \in B_n} (1 + a_Sh_S)\, dy
		\end{equation}
		where $a_S \in [-1,1]$ and $a_S$ is the product coefficient for $S.$

		\item Any assignment of parameters $a_S$ for $(-1,1)$ and choice of $\mu(X)>0$ determines a measure $\mu$ which is positive on all sets $S$ on $B$ with product formula
		${\mu=\mu(X) \prod_{S \in B_n} (1 + a_Sh_S)\, dy}$
		whose product coefficients are the parameters $a_S.$
		
		\item Any assignment of parameters $a_S$ from $[-1,1]$ and choice of $\mu(X)>0$ determines a non-negative measure $\mu$ with product formula ${\mu=\mu(X) \prod_{S \in B_n} (1 + a_Sh_S)\, dy}.$ The parameters are the product coefficients if they satisfy the constraints:
		\begin{enumerate}
			
			\item If $a_S=1$, then the product coefficient for the tree rooted at $R(S)$ equals 0.
			\item If $a_S=-1$, then the product coefficient for the tree rooted at $L(S)$ equals 0.
		\end{enumerate}

	\end{enumerate}

\end{lemma}

\subsection{Real world applications}\label{subsec: applications}
 Advancement on this project will be of great
impact in studying climate change. One of the most important structural properties of vegetation (leaf area index) is related to the rate at which forests grow and sequester carbon which is also an important factor for micro climates within
forests that may help mitigate some of the initial impacts of climate change.

Differentiation of photosynthetic components (leaf, bushes or grasses) and non-photosynthetic components (branches or stems) by
3D terrestrial laser scanners (TLS) is of key importance to understanding the spatial distribution of the radiation
regime, photosynthetic processes, and carbon and water exchanges of the forest canopy. Research suggests that
woody canopy components are a major source of error in indirect leaf area index (LAI) estimates. We expect that
the use of a deep learning framework will improve the accuracy of the quantification of woody material, hence
improving the accuracy of LAI estimates (see \cite{Zheng2009} for background).

\subsection{Computing product coefficients on Lidar}\label{subsec: PCs on lidar}

The main purpose of this section is to show how to compute the product coefficients defined in Sec.\,\ref{PCs}

We apply the product formula representation from Lemma \ref{lemma1} to a counting measure derived from our 3D LiDAR point cloud dataset. Each point in the dataset is assigned a classification label corresponding to high vegetation, ground, building, or low vegetation.  

Prior studies, such as \cite{BaIzMcNeSh:2012}, explored this dataset using a multi-scale Singular Value Decomposition (SVD) approach to develop an SVM-based classification model. Their results demonstrated that vegetation and ground points could be accurately distinguished using this technique. More recently, \cite{Linda} proposed an alternative method that utilized product coefficient parameters as classification features instead of multi-scale SVD descriptors. However, that analysis was limited to a binary classification task.  

In contrast, our work extends these approaches by performing multi-class classification while incorporating product coefficients computed locally on a per-point basis. To evaluate the effectiveness of these additional features, we apply K-Nearest Neighbors (KNN) and Random Forest classifiers and assess how classification performance improves after applying Principal Component Analysis (PCA) to the newly generated feature set.  

The findings from \cite{BaIzMcNeSh:2012} suggest that decision rules for distinguishing between two distributions—such as vegetation versus ground—can be inferred from histograms of product coefficients. Although the classification performance was not as strong as that achieved with multi-scale SVD, the approach provided a more interpretable rationale for the decision boundaries.

\begin{figure}\label{fig:binary}
\centering	
\includegraphics[scale=0.6]{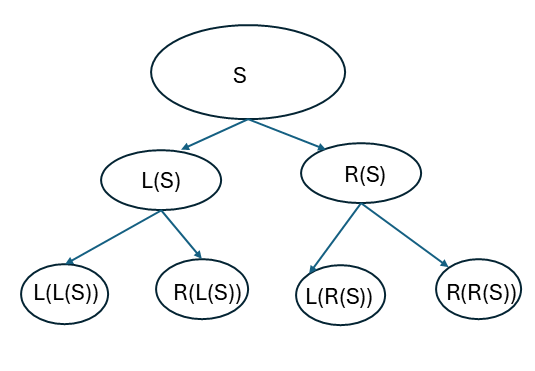}
\caption{The diagram represents the binary tree for a general set $S.$ The diagram includes the notation used in the computation of the seven product coefficients. There are $2^0, 2^1$ and $2^2$ product coefficients per level. }

\end{figure}

\begin{figure}
\includegraphics[scale=0.5]{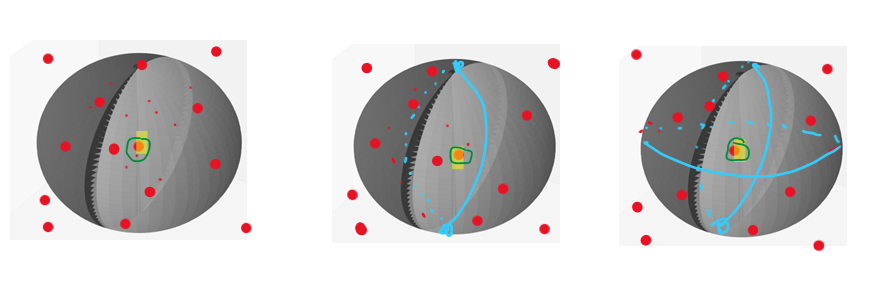}
\caption{When computing product coefficients per point $(x_i, y_i, z_i)$, consider a sphere $S+i$of radius 2. We first slice the sphere in along the $x$-axis and compute the first product coefficient $a_S$, then slice the sphere along the $y-$ axis and compute two product coefficients. One for the left child $L(S_i)$ and the other for the right child $R(S_i).$ Last, we slice along the $z-$ axis and compute the the last four product coefficients.} 
\label{fig:spheres}
\end{figure}

The general procedure to compute product coefficients goes as follow:

\begin{enumerate}
\item Normalize the original 3D point cloud to fit on the unit cube ${[0,1]}^3.$ Count the points of the Lidar dataset $S.$

\item For each fixed data point $(x_i,y_i,z_i)$, consider a sphere $S_i$ of radius 2.  Slice the sphere along the $x-$ axis direction and compute the product coefficient $a_S$ by counting the number of points in the left child $L(S)$ and the right child $R(S_i)$. The left child, $L(S_i)$ consists on all the points on the left section of the sphere and the right child the number of points in the right section of the sphere. 

\item Slice the sphere in the direction of the $y-$ axis and compute the product coefficients $a_{L(S_i)}$ and $a_{R(S_i)}.$ Notice that now the initial parent set is $L(S_i)$ or $R(S_i)$.

\item Finally, slice the sphere along the $z-$axis and compute the product coefficients $a_{L(L(S_i)},\, a_{R(L(S_i))},\, a_{L(R(S_i))}$ and $a_{R(R(_iS))}.$ See Fig,.\,\ref{fig:binary} for illustration of the binary tree and the notation for the children at each level. 
\end{enumerate}
This above procedure is shown in Fig.\,\ref{fig:spheres}. We have a total of seven product coefficients for each point $(x_i,y_i,z_i$).

We end up with seven new features and add them to the original spatial features $x,y$ and $z.$ 

\subsection{Classification experiments}
\label{subsec: PCA and PCs experients}

This section aims to examine the performance differences between two classification algorithms, K-Nearest Neighbors (KNN, with \( k=10 \)) and Random Forest (with 100 trees), in classifying the four LiDAR-derived classes in our original dataset. We compare this with classification based on a set of seven newly derived features obtained from two levels of product coefficients computed within a local neighborhood. This neighborhood is defined by a sphere of radius 2 centered at each data point, structured in a dyadic tree, and utilizing a counting measure with Principal Component Analysis (PCA) applied to retain up to three principal components.

The Nearest Neighbor classifier, particularly its K-Nearest Neighbors variant, is ideal for geospatial data as it utilizes spatial relationships between data points for classification and prediction. It determines the class or value of a new data point based on its proximity to neighboring points. On the other hand, Random Forest classifiers are highly effective for geospatial data analysis due to their ability to handle complex, high-dimensional datasets, capture non-linear relationships, and offer insights into feature importance.

Our proposed methodology offers several advantages. First, since dimensionality reduction is applied without using the data labels, this step minimizes the risk of overfitting. Second, while the original classifiers do not perform well on the base dataset, adding product coefficient features from localized neighborhoods and applying PCA with multiple principal components significantly enhances classification accuracy.

We conducted two main sets of experiments. In the first set, classification was performed using only the spatial coordinates (x, y, and z) as features, with both KNN and Random Forest classifiers. We then compared these results with classifications that included the seven new product coefficient features computed in the local neighborhood. In the second set, we applied the same classifiers using the generated seven features and implemented PCA to extract between three and ten principal components, which were used as inputs to the classifiers.


The main steps of our proposed algorithm, incorporating product coefficients as features along with Principal Component Analysis (PCA), are as follows:

\begin{enumerate}
	\item \textbf{Feature Generation with Product Coefficients:} Starting with the original dataset, we generate a new feature set $X$ by computing product coefficients within localized neighborhoods. Specifically, we define a ball of radius 2 centered at each data point and use the points within this ball as a "parent" set for calculating the first-level product coefficient. In the subsequent level, two product coefficients are computed—one for the left child and one for the right child of this dyadic tree structure. At the following level, we compute $2^2$ additional product coefficients. This process yields seven new features for each data point. These features are then normalized to fit within the unit cube.
	
	\item \textbf{Dimensionality Reduction using PCA:} We apply PCA to the newly generated input data $X$ to create a transformed input layer $Z$ of varying dimensions, specifically with 3 to 10 principal components. This results in eight different experimental setups, each with a unique number of principal components.
	
	\item \textbf{Classification with KNN and Random Forest:} The transformed input $Z = XM \in \mathbb{R}^{n}$, where $M$ is the matrix of principal components derived from the covariance matrix of the input data, is then used as input to two classifiers—K-Nearest Neighbors (KNN) and Random Forest. Each experiment evaluates classifier performance as the dimensionality of $Z$ increases from $n=3$ to $n=10$. The cross-validated F1 scores for each setup are presented in Table~\ref{table:F1scoresPCsPCA}.
\end{enumerate}

This structured approach allows us to evaluate the impact of localized product coefficients and PCA-based dimensionality reduction on classification accuracy.

\begin{table}
	\centering	
	\begin{tabular}{c | c | c}
		\hline	
		& KNN $F^1$-score & RF  $F^1$-score   \\
		\hline         
		Original features $(x.y,z)$  & 0.33 ($\pm$ 0.18)       &   0.41 ($\pm$ 0.16)  \\
		
		\hline
		With PCs        &0.33  ($\pm$ 0.18)           & 0.45   ($\pm$ 0.15)                \\
		\hline

	\end{tabular}
	\caption{cross validation on $F_1-$ scores using KNN and random forest (RF) on the original data with only $x,y,$ and $z$ spatial coordinates as features and the data with the new seven product coefficient generated features}
	\label{table:F1scoresNoPCvsPC}
\end{table}
%

\begin{table}
\centering
\centering
\begin{tabular}{c | c | c}
		\hline	
		 \# of Principal Components & KNN $F^1$-score & RF  $F^1$-score   \\
		
		\hline
		     3           & 0.39 ($\pm$ 0.12) &  0.38 ($\pm$ 0.12)  \\
		      4           &0.45 ($\pm$ 0.10)  &   0.39 ($\pm$ 0.10)  \\
		       5          & 0.48 ($\pm$ 0.11)     & 0.38 ($\pm$ 0.16)  \\
		       6           &0.56 ($\pm$ 0.08)   &  0.44 ($\pm$ 0.15)\\
		       7           & 0.61 ($\pm$ 0.06)   &  0.49 ($\pm$ 0.14) \\
		       8           & 0.65 ($\pm$ 0.05)   &  0.55 ($\pm$ 0.14) \\
		       9           & 0.67 ($\pm$ 0.04)    &  0.52 ($\pm$  0.14) \\
		      10          & 0.85 ($\pm$ 0.02)    &  0.81 ($\pm$ 0.16)  \\
		\hline

	\end{tabular}
\caption{$F^1$ scores using PCA on the data with the new seven product coefficient generated features. Cross validation on $F_1-$ scores using $n=3,\ldots,10$ principal components.}
\label{table:F1scoresPCsPCA}
%
%
%
%
\end{table}

First observe the $F_1$ scores in Table\,\ref{table:F1scoresNoPCvsPC}. The new generated features don't improve the accuracy when using KNN without PCA. However, we can see that the best accuracy of 0.45 is achieved by RF when using the new product coefficient features. 

A big improvement is observed whenever we use the the new generated product coefficients combined with PCA, as seen on Table\,\ref{table:F1scoresPCsPCA}. Note that the accuracy of both classifiers improves as we go from three principal components up to ten components. Going from 0.39 ($\pm$ standard deviation) to 0.85. We attribute this to the fact that PCA is minimizing the covariance. Also, KNN performs slightly better than RF which might be a consequence of working with geospatial data.

In this paper, we did not analyze computational cost as the dataset was relatively small, and our primary focus was on improving accuracy. In future work, we plan to investigate computational efficiency on larger datasets.

\section{Conclusion and Future Research Directions}
\label{summary}
In this paper, we showed that using quantities such as product coefficients can enrich the original 3D point cloud Lidar dataset by computing seven product coefficients on dyadic trees on local neighborhoods for each data point. We performed experiments using two classic machine learning classifiers and observed that when combined with a PCA with a high number of components, the performance of the classifiers improves. 

The results of our experiments demonstrate that the dimensionality reduction technique employed effectively reduces covariance among features, thereby improving classifier performance. However, it is important to note that other dimensionality reduction techniques might perform equally well or even better under different conditions. Future work will explore alternative methods to further enhance performance and assess their suitability for geospatial data classification.

This study does not address computational cost, as the dataset used is relatively small, with our focus centered on enhancing accuracy. Future research will explore computational efficiency using larger datasets, such as the Golden Gate Bridge LiDAR dataset containing 15 million points.

Future directions include: 
\begin{itemize}
\item considering more physical features from the original Lidar and computing product coefficients on neighborhood sets of dimension $n \geq 3.$ 

\item considering bigger size data sets to test the robustness of this new framework.

\item engineering more features by computing intrisic dimension for example of the dataset and perform more comparison experiments. 

\item performing dimensionality reduction with an auto-encoder and comparing this new framework with the one including PCA. 

\item studying other covariance reduction techniques and compare with PCA performance.

\end{itemize}

\begin{acknowledgement}
I want to thank Dr. Linda Ness for introducing me to product coefficients in Lidar and to Dr. Boris Iskra from AdTheorend Inc for helping me with debugging my code. Also, I want to thank Dr. Rasika Karkare for the initial dicussions on code implementation.  
\end{acknowledgement}


\bibliographystyle{siam} 
\bibliography{IntegratingProductCoefficientsforImproved3DLiDARDataClassificationL}
\end{document}